\documentclass[10pt,twocolumn,letterpaper]{article}

\usepackage{cvpr}
\usepackage{times}
\usepackage{epsfig}
\usepackage{graphicx}
\usepackage{amsmath}
\usepackage{amssymb}
\usepackage{subcaption}
\usepackage{booktabs}
\usepackage{color}
\usepackage[pagebackref=true,breaklinks=true,letterpaper=true,colorlinks,bookmarks=false]{hyperref}

\cvprfinalcopy 

\def\cvprPaperID{5993} 

\newcommand{\hk}[1]{{\color{black}{#1}}}
\newcommand{\hknew}[1]{{\color{black}{#1}}}

\begin{document}

\title{ Dynamics are Important for the Recognition of Equine Pain in Video}


\author{\parbox{16cm}{\centering
    {\large Sofia Broom{\'e}$^1$ ~~~ Karina Bech Gleerup$^2$ ~~~ Pia Haubro Andersen$^3$ ~~~ Hedvig Kjellstr{\"o}m$^1$}\\
    {\normalsize
$^1$ Division of Robotics, Perception, and Learning, KTH Royal Institute of Technology, Sweden\\
$^2$ Section for Medicine and Surgery, University of Copenhagen, Denmark\\
$^3$ Deptartment of Clinical Sciences, Swedish University of Agricultural Sciences, Sweden\\ {\tt\small sbroome@kth.se ~~ kbg@sund.ku.dk ~~ pia.haubro.andersen@slu.se ~~ hedvig@kth.se}}}}

\maketitle

\begin{abstract}

        A prerequisite to successfully alleviate pain in animals is to recognize it, which is a great challenge in non-verbal species. Furthermore, prey animals such as horses tend to hide their pain. In this study, we propose a deep recurrent two-stream architecture for the task of distinguishing pain from non-pain in videos of horses. Different models are evaluated on a unique dataset showing horses under controlled trials with moderate pain induction, which has been presented in earlier work. Sequential models are experimentally compared to single-frame models, showing the importance of the temporal dimension of the data, and are benchmarked against a veterinary expert classification of the data. We additionally perform baseline comparisons with generalized versions of state-of-the-art human pain recognition methods. While equine pain detection in machine learning is a novel field, our results surpass veterinary expert performance and outperform pain detection results reported for other larger non-human species.
    
\end{abstract}

\section{Introduction}

\hk{This paper presents a method for automatic detection of equine pain behavior in video} -- a highly challenging problem \hknew{since horses display pain through} subtle 
signals.

Recognition of pain in animals is important because pain compromises animal welfare and can be a manifestation of disease. Pain diagnostics for humans typically include self-evaluation with the help of standardized forms and labeling of the pain by a clinical expert using pain scales. However, animals cannot verbalize their pain. The use of standardized pain scales is moreover challenged by the fact that prey animals such as horses and cattle display subtle and less obvious pain behavior. It is simply beneficial for a prey animal to appear healthy, to lower the interest from predators.

Pain in horses is instead typically assessed \hk{by (manually)} observing a combination of behavioral traits such as exploratory behavior, restlessness, positioning in the box and changes in facial expressions \cite{Wathan2015}. Video filming may be used to prolong the observation periods and increase the likelihood of capturing pain behavior. However, proper manual annotation such as EquiFACS \cite{Wathan2015} of the facial expressions and other behavioral signs of pain is time consuming \cite{McDonnell2005} -- it can take two hours to evaluate a two minute video clip. This severely limits the possibilities for routine pain diagnostics on horses, for instance before or after surgery. An automatic pain detection system would allow veterinarians and horse owners to reliably and frequently screen the pain level of the horse. Section \ref{paininhorses} provides more background on equine pain from a veterinary point of view.

We introduce a new research direction and application field in computer vision: detection of equine pain in video. While human pain detection has been performed with single images, owing to our expressive faces, \hk{the} research hypothesis \hk{in this paper is} that the pose and movement patterns matter more for horses. The equine pain detection task is explored using recurrent neural network architectures. These memory-preserving networks are chosen in order to model temporality, which is considered crucial when assessing pain in horses \cite{Wathan2015}. We have not seen previous work in pain detection for non-human species where temporal information is taken into account.

\hk{Specifically, we propose a deep neural network approach to automatic detection of horse pain in video, taking} 
both spatial and temporal context into account (Section \ref{sec:models}). Our architecture is a new combination of existing network types, using optical flow as an attention mask in a Convolutional LSTM \cite{Shi} two-stream setting. We have reinterpreted the two-stream network first presented by \cite{Simonyan2014} into a fully recurrent two-stream Convolutional LSTM network. The fusion of the two streams is done on the feature-level, adding the optical flow feature map to the RGB feature map element-wise, which highlights the regions of interest \hk{in the video}. 
The method learns patterns end-to-end without the help of intermediate representations such as the Facial Action Coding System (FACS) \cite{ekman1976measuring}. To the best of our knowledge, this has not been done previously for animals.

\hk{The method is experimentally shown to outperform several baselines, including a method relying on single-frame analysis. Most notably,
we show that our method performs better than manual classification by veterinarians with extensive expertize in equine pain assessment.}



\section{Background and related work}
\label{relwork}
\subsection{Pain evaluation in horses}
\label{paininhorses}
Within the last decade, there has been an increasing focus on pain evaluation in horses, both because there is an increasing knowledge about the detrimental effects of pain and because advanced treatments and surgical procedures have become more common for horses \cite{GleerupLindegard2016}. Compared to small animals, horses are often not given the optimal pain treatment \cite{Love2009}. Reducing pain to a minimum ensures animal welfare, improves convalescence and optimizes treatment success \cite{Sellon2004}. Pain detection and quantification in horses depend on an observer to detect potentially pain related changes in behavior and in physiological parameters.

There is increasing focus on behavioral observations, since physiological parameters are often not pain-specific \cite{Bussieres2008}. When in pain, horses change their activity budget \cite{Price2003,Pritchett2003}, meaning the percentage of their time that they spend lying down, eating, being attentive, etc. However, documenting this is very time consuming. Even though short, applicable pain scoring tools have been
developed \cite{GleerupLindegard2016}, these do not capture all pain behavior but rather do spot sampling. An automated means to record pain behavior from surveillance videos would be an important step forward in equine pain research.

Facial expressions have lately been presented as a sensitive measure of pain in horses, when observed in combination with other pain behaviors \cite{DallaCosta2014,Gleerup2015,GleerupLindegard2016,vanLoon2015}. A few examples of the facial pain characteristics described in \cite{Gleerup2015} are slightly dropped ears, muscle tension around the eyes with an increased incidence of exposing the white in the eye, a square-like shape of the nostrils and an increased tension of the lips and chin.

\subsection{Automated pain recognition}
\label{emorecog}
Automated pain recognition has largely focused on human pain. Littlewort et al.~\cite{Littlewort1844} distinguish between posed and real pain for humans in video by training separate linear support vector machine (SVM) classifiers for different FACS \cite{ekman1976measuring} action units. Among the works on pain recognition in humans using deep learning are Kahou et al.~\cite{EbrahimiKahou2015a}, Rodriguez et al.~\cite{Rodriguez2017} and Zhou et al.~\cite{Zhou2016}. \cite{Rodriguez2017} use an LSTM layer on top of a VGG-16 \cite{VGG16paper} CNN. 
Their task is to detect pain in the UNBC-McMaster Shoulder Pain Database \cite{LuceyPainfulData2011}, which is a video dataset for human pain recognition. Features up until and including the first dense layer in VGG-16 are extracted frame-wise and then arranged in sequences that are fed to the LSTM. The top LSTM layer is found to improve the result significantly compared to the base model. By aligning, cropping, masking and frontalizing the input data to the VGG+LSTM architecture they obtain the best known result on this dataset. Albeit without the pre-processing, we have included tests of their architecture on our dataset in Section~\ref{sec:results}.

Lu et al.~\cite{LuMahmoud} explore automated pain assessment in sheep faces from still images. To our knowledge, this is the only previous work in automatic recognition of pain for larger animals. Facial action units associated with pain
are detected
through the successive steps of facial detection, landmark detection, feature extraction and classification with an SVM. Our system performs better than the one of \cite{LuMahmoud} on the binary pain detection task for unseen subjects.

Tuttle et al.~\cite{MiceNN} use deep learning to perform automatic facial action unit detection in mice, to score for pain. The single-frame InceptionV3 CNN is trained to detect action units, which are counted according to a grimace scale to establish the presence of pain.

Differently from the work of \cite{LuMahmoud} and \cite{MiceNN}, our method takes an end-to-end approach to the problem, so as to not depend on pre-defined pain cues. We do not only study the face but also the pose, body and movement pattern of the animal, which we believe is crucial for reliable pain diagnostics in horses. The data both in \cite{LuMahmoud} and in \cite{MiceNN} are labeled according to a grimace scale ground truth and not according to whether pain was induced or not. Furthermore, the respective datasets are both culled so that only clear frontal faces are considered. 

\subsection{Action recognition in video}
\label{actionrecognition}
Action recognition methods can guide model development for pain recognition in video data, since both fields require the capability of extracting information from sequences of images.
The challenge in action recognition and computer vision for video lies in fully using the information captured in dependencies across the temporal dimension of the data. The task of recognizing movements and gestures from video is still far from human performance, in contrast to single image object recognition. There are different ways of utilizing temporal information across a sequence of images. In a neural network context, this can for example be achieved using 3D convolutional layers, pooling across the temporal dimension, or by using recurrence. Long Short-Term Memory (LSTM) networks \cite{LSTM1997} are Recurrent Neural Networks (RNNs), proven to work well for data with long-term dependencies since they can mitigate the problem of vanishing gradients, from which standard RNNs suffer \cite{OnDifficultyTrainRNN}. 

An early work in action recognition using deep learning by Karpathy et al.~\cite{Karpathy2014} compares single-frame and sequential models and finds only a marginal difference between the two approaches. But more recently, sequential models have clearly outperformed the ones that are trained on single frames \cite{QuoVadis2017, JoeYue-HeiNgMatthewHausknechtSudheendraVijayanarasimhanOriolVinyalsRajatMonga2015, Wang2016a}. Single-frame models base their classification on image content instead of on the dynamics between frames. This works to some extent, especially on datasets such as Sports-1M \cite{Karpathy2014} or UCF-101 \cite{UCF-101} where most of the activity categories contain pertinent visual objects -- consider for example walking a dog or playing the piano. 
 
Ng et al.~\cite{JoeYue-HeiNgMatthewHausknechtSudheendraVijayanarasimhanOriolVinyalsRajatMonga2015} mainly test two approaches to the sequential activity recognition problem, after deep two-stream CNN \cite{Simonyan2014} feature extraction of the frames and their corresponding optical flow.
Stacked LSTM layers are compared to different ways of feature pooling across the temporal dimension. The former approach obtained the thereto highest result on the Sports-1M dataset.

Concerning the two-stream model, in \cite{EbrahimiKahou2015a} a comparison is made between stream fusion at the feature-level vs. fusion at the decision-level (meaning typically after the softmax output), and the authors find that feature-level fusion improves their results. In our two-stream architecture experiments, we follow their example and fuse the streams at the feature-level.

Shi et al.~\cite{Shi} introduce the Convolutional LSTM (C-LSTM) unit, where the fully-connected matrix multiplications involving the weight matrices in the LSTM equations \cite{LSTM1997} are replaced with convolutions. The idea with the C-LSTM is to preserve spatial structure and avoid having to collapse the input images to vectors, as one needs to do for a standard LSTM (Section \ref{sec:clstm}). Networks composed of C-LSTM units were the best-performing architectures in our experiments on the equine dataset.

\begin{table}[!b]
\vspace{-3mm}
\setlength{\tabcolsep}{5pt}
\centering
\caption{\small Dataset overview, per horse subject and pain label (hh:mm:ss).}
\vspace{-2mm}
\label{horsetimes}
\begin{tabular}{@{}l|llllll|l@{}}
\toprule
\small Subj. ID & \small 1        & \small 2        & \small 3        & \small 4        & \small 5        & \small 6        & \small Total    \\ \hline
\scriptsize Gender & \scriptsize Mare        & \scriptsize Mare        & \scriptsize Mare        & \scriptsize Mare        & \scriptsize Mare        & \scriptsize Gelding        & \scriptsize n/a   \\
\scriptsize Age (years) & \footnotesize 14        & \footnotesize 7        & \footnotesize 12        & \footnotesize 6        & \footnotesize 14        & \footnotesize 3        & \footnotesize n/a   \\ \hline
\scriptsize Pain     & \tiny 00:58:55 & \tiny 00:39:30 & \tiny 00:29:52 & \tiny 00:22:21 & \tiny 00:44:38 & \tiny 00:26:09 & \tiny 03:41:25 \\
\scriptsize No pain  & \tiny 01:00:27 & \tiny 01:01:29 & \tiny 01:01:10 & \tiny 01:31:12 & \tiny 00:59:49 & \tiny 00:29:37 & \tiny 06:03:44 \\ \hline
\scriptsize Total    & \tiny 01:59:22 & \tiny 01:40:59 & \tiny 01:31:02 & \tiny 01:53:33 & \tiny 01:44:27 & \tiny 00:55:46 & \tiny 09:45:09 \\
\tiny \# frames, 2 fps    & \scriptsize 14324 & \scriptsize 12118 & \scriptsize 10924 & \scriptsize 13626 & \scriptsize 12534 & \scriptsize 6692 & \scriptsize 70292 \\\bottomrule
\end{tabular}
\end{table}

The C-LSTM in \cite{Shi} is applied to weather prediction with radar map sequences as input. But owing to the contribution of \cite{Shi}, C-LSTM layers have recently been used in a few works on action and gesture recognition from video data, notably by Li et al.~\cite{LiVideoLSTM2016} and Zhu et al.~\cite{ZhuCLSTM2017}. The architecture of \cite{ZhuCLSTM2017} consists of two streams: one using RGB and one using RGB-D as input. The streams \textcolor{black}{are each} composed of a 4-layer 3D CNN followed by a 2-layer C-LSTM network, spatial pyramid pooling and a dense layer before fusion. Similar to our method, \cite{ZhuCLSTM2017} perform simultaneous extraction of temporal and spatial features. Their idea is to let the 3D convolutions handle the short-term temporal dependencies of the data and the C-LSTM the longer-term dependencies. We differ from the method of \cite{ZhuCLSTM2017} in that we use optical flow and not RGB-D data for the motion stream and that our parallel streams consist only of recurrent C-LSTM layers, without 3D convolutions.

Li et al.~\cite{LiVideoLSTM2016} introduce the VideoLSTM architecture, using C-LSTM networks in combination with a soft attention mechanism both for activity recognition and localization. Features are first extracted from RGB and optical flow frames using a single-frame VGG-16 \cite{VGG16paper} pre-trained on ImageNet \cite{Deng} before they are fed to the recurrent C-LSTM streams. The attention in \cite{LiVideoLSTM2016} is computed by convolving the input feature map at $t$ with the hidden state from $t-1$. A given C-LSTM stream in their pipeline always has one layer, consisting of 512 units. By contrast, our C-LSTM network is deep but has fewer units per layer, and we do not separate feature extraction from the temporal modeling.

\vspace{-1mm}


\section{The Equine Pain Dataset}
\label{sec:dataset}


\hk{The equine pain dataset used in this paper has been collected by Gleerup et al.~\cite{Gleerup2015}, and} consists of 9 hours and 45 minutes of video across six horse subjects, out of which 3 hours and 41 minutes are labeled as pain and 6 hours and 3 minutes as non-pain. These binary labels have been set according to the presence of pain induction, known from the recording protocol (Section \ref{pain}). Frames are extracted from the equine videos at 2 fps. When training on 10-frame sequences, the dataset contains $\sim 7$k sequences. Duration-wise this dataset is comparable to known video datasets such as Hollywood2 \cite{marszalek09} (20 hours) and UCF-101 \cite{UCF-101} (30 hours). Details on the class distribution across the different subjects can be seen in Table \ref{horsetimes}. A more complete description with image examples can be found in \cite{Gleerup2015}.

\subsection{Recording setup and pain induction}
\label{pain}

The subjects were trained with positive reinforcement before the recordings to be able to stand relatively still in the trial area. They were filmed during positive and negative pain induction trials with a stationary camera at a distance of approximately two meters. The imagery contains the head and the body until the wither.
The camera was fixed, which means that the data is suited for extraction of optical flow, since there is no camera motion bias \cite{Karpathy2014}.

Pain was induced using one out of two noxious stimuli applied to the horses for 20 minutes: a pneumatic blood pressure cuff placed around the antebrachium or the application of capsaicin 10\% (chili extract) on 10 $\text{cm}^2$ skin \cite{Gleerup2015}. Both types of experimental pain are ethically regulated and also occur in human pain research. They caused moderate but completely reversible pain to the horses \cite{Gleerup2015}.

\subsection{Noise and variability}
\label{sect:noisydata}

The videos in the dataset present some challenging noise. The lighting conditions were not ideal and some videos are quite dark. Unexpected elements like a veterinarian standing by the horse for a while, frequent occlusion of the horse's head, varying backgrounds, poses, camera angles and colors on halters all contribute to the challenging nature of the dataset. The coat color of the horses is mostly dark brown except for one chestnut but they do have varying characteristics such as stars and blazes.

Furthermore, there is some variability among the subjects. Subject 6 is by far the youngest horse at age 3, and frequently moves in and out of the frame. This presented difficulties for our classifiers as can be seen in Table \ref{table:persubject}. Importantly, this subject did not go through the same training for standing still as the other subjects. We include test runs on this horse for transparency.


\section{Approach}
\label{sec:models}

Three main architectures are investigated, and each is additionally extended to a two-stream version. With two streams, we can feed sequences of different modalities to the network. In our case, we use RGB in one stream and optical flow in the other. The optical flow of two adjacent frames is an approximation of the gradients ($u$ and $v$) of the pixel trajectories in the horizontal and vertical directions, respectively. We compute the optical flow using the algorithm presented by \cite{Farneb2003} and add the magnitude of $u$ and $v$ as a third channel to the tensor. The one-stream models receive either single frames or sequences of either modality.

The three base architectures are the deep CNN InceptionV3 \cite{Szegedy2014}, which takes single-frame input, the partly recurrent VGG+LSTM architecture from \cite{Rodriguez2017} which is a VGG-16 CNN \cite{VGG16paper} up to and including the first dense layer with one LSTM layer on top, and the fully recurrent C-LSTM. Intuitively, the three models differ in the extent by which they can model the dynamics of the data. InceptionV3 is a static model that only learns patterns from single frames. In its two-stream setting, it processes single optical flow frames in parallel to the RGB frames, which adds limited motion information. The VGG+LSTM model extracts temporal features, but does so separately from the spatial features. Its recurrent top layer is not an integral part of the network, as it is for the C-LSTM model. In the C-LSTM architecture all layers are fully recurrent and the spatial and temporal feature extraction take place simultaneously.

\subsection{Convolutional LSTM}
\label{sec:clstm}

Shi et al.~\cite{Shi} introduce the Convolutional LSTM (C-LSTM) unit, where the fully-connected matrix multiplications involving for instance the input-hidden and hidden-hidden weight matrices in the LSTM equations \cite{LSTM1997} are replaced with convolutions. This enables the parameter sharing and location invariance that comes with convolutional layers, while maintaining a recurrent setting. This type of layer structure is not to be confused with the more common setup of stacking RNN layers on top of convolutional layers (as in the VGG+LSTM architecture), which differs in two important and related aspects from the C-LSTM. First, any input to a standard RNN layer has to be flattened to a one-dimensional vector beforehand, which shatters the spatial grid patterns of an image. Second, by extracting the spatial features separately, hence by performing down-sampling convolutions of the images before any sequential processing, one risks losing important spatio-temporal features. In effect, there are features of a video clip that do not stem exclusively from either spatial or temporal data, but rather from both. For this reason, C-LSTM layers are especially suited to use for video analysis.

Next, we present the details of the C-LSTM networks in one and two streams from our experiments.

\vspace{-4mm}
\paragraph{Convolutional LSTM in one and two streams.}
 The one-stream C-LSTM model (C-LSTM-1) has four stacked layers of 32 C-LSTM units each with max pooling and batch normalization between every layer, followed by a dense layer and sigmoid output. 

\begin{figure}[b!]
\vspace{-3mm}
\centerline{\includegraphics[width=1\linewidth]{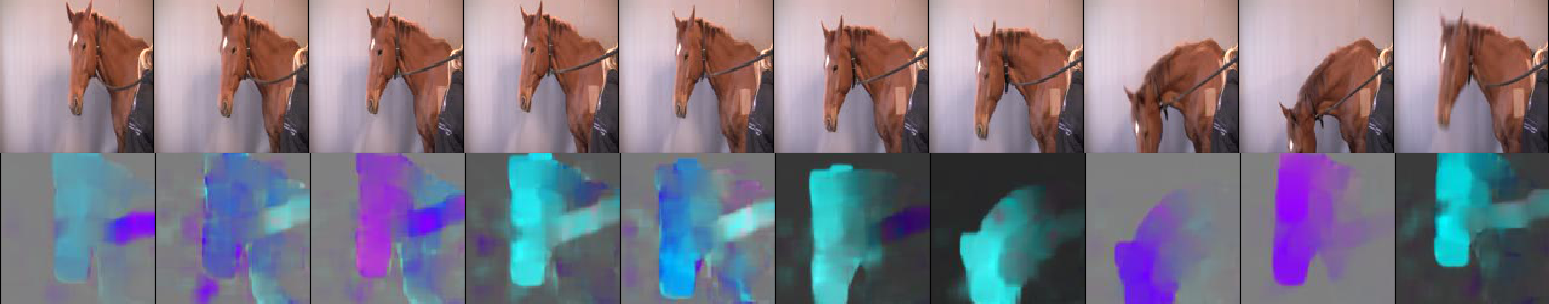}}
\vspace{-2mm}
\caption{\small Example sequence from the dataset. Optical flow on second row. In the two-stream model, the optical flow serves as attention when fused with the RGB frames.}
\label{fig:exampleseq_OF}
\end{figure}

\begin{figure*}[t!]
\centerline{\includegraphics[width=1\linewidth]{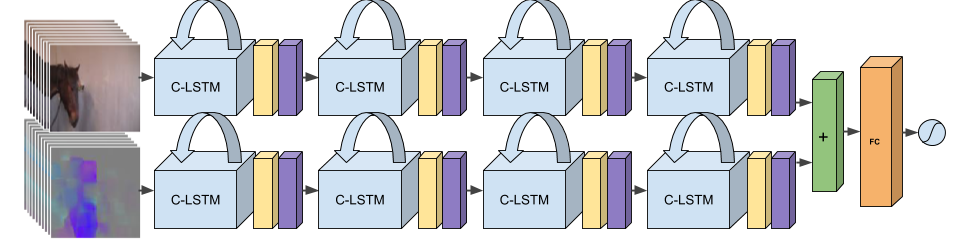}}
\vspace{-2mm}
\caption{\small Our fully recurrent C-LSTM-2 model, which performed the best on the equine dataset. The yellow and purple layers are max pooling and batch normalization, respectively. The arrows on top of the C-LSTM layers symbolize recurrence.}
\vspace{-3mm}
\label{fig:clstm2}
\end{figure*}

The two-stream C-LSTM (C-LSTM-2) is shown in Figure~\ref{fig:clstm2}. This model consists of two parallel C-LSTM-1 streams. The idea, just like in \cite{Simonyan2014}, is for the motion stream (optical flow) to complete the spatial stream (RGB). However, our C-LSTM two-stream model differs in some aspects from the original architecture presented in \cite{Simonyan2014}. To begin with, in \cite{Simonyan2014}, only the motion stream takes input with a temporal span while the spatial one processes single (momentaneous) RGB frames. Furthermore, the motion stream input, consisting of $k$ concatenated optical flow frames, is not treated recurrently but as one single input volume. In \cite{Simonyan2014}, both streams are feedforward CNNs, whereas in our case both streams are recurrent.
In addition, we choose to fuse our two streams at the feature-level, in contrast to \cite{Simonyan2014} who fuse the class scores only after the softmax output layer. In particular, we fuse the two tensor outputs from the fourth layers in both streams by either element-wise multiplication or addition (Table \ref{table:1streamresults}), in order to use the optical flow features as an attention mask on the RGB features. To this end, the optical flow is computed using rather large averaging windows, which results in soft, blurry motion patterns, similar to the attention results presented in \cite{Sharma2015} (Figure~\ref{fig:exampleseq_OF}). When using multiplication, the attention mask behaves like an $\text{AND}$ gate; when using addition, it behaves like the softer $\text{OR}$ gate. Both types of fusion emphasize the parts of the image where pixels have moved, which are likely to be of interest to the model.

\subsection{Implementation details}
The top dense and LSTM layers of the VGG+LSTM model are trained from scratch and the convolutional layers of the VGG-16 base model are pre-trained on ImageNet \cite{Deng} as an initialization. Its LSTM layer has 512 units, which hyperparameter was set after trials with [32, 256, 512, 1024] units. InceptionV3 is trained from scratch since this worked better than with ImageNet pre-trained weights. The top part of the original model is replaced with a dense layer with 512 units and global average pooling before the output layer.
All architectures have a dense output layer with sigmoid activation and binary cross-entropy as loss function. The respective optimizers used are shown in Table \ref{table:impl}. For the VGG+LSTM, we also trained with the Adam optimizer, as in \cite{Rodriguez2017}, but Adadelta gave better results.
We use larger frames when training the InceptionV3 models than when training the sequential models (Table \ref{table:impl}) because InceptionV3 has a minimum input size. When training on sequences, they consist of 10 frames each. The sequences are extracted back-to-back without overlap (stride 10). We use early stopping of 15 epochs on a maximum of 100 epochs throughout the experiments. All two-stream models use dropout with probability 0.2 after the fusion of the two streams. Since additive feature level fusing worked best for the C-LSTM-2, this is chosen as the method of fusion for the two-stream extensions of VGG+LSTM and InceptionV3 as well. Data augmentation for the sequences consists of horizontal flipping, random cropping and shading by adding Gaussian noise. The transformations are consistent across a given sequence (i.e. for random cropping, the same crop is applied for the whole sequence). Remaining hyperparameters and the code for the experiments can be found at \url{https://goo.gl/9TPYNk}.


\section{Experiments and results}
\label{sec:results}

\subsection{Evaluation method}

The classification task is evaluated using leave-one-subject-out cross testing, meaning that for the equine dataset we always train on four subjects, validate on one and test on the remaining subject. By separating the subjects used for training, validation and testing respectively, we enforce generalization to unseen subjects.
Subject 5 is always used as validation set for early stopping, on account of its relatively even class balance (Table \ref{horsetimes}). When 
it is tested on, subject 1 is used for validation instead. The results presented are the average of the six cross-subject test rotations, which additionally were repeated five times. The standard deviations in Table \ref{table:1streamresults} are the averages of the standard deviations across the test folds for every run, while the standard deviations presented in Table \ref{table:persubject} are the per-subject variations across the five runs, which are smaller. F1-score is the harmonic mean of precision and recall. The reason to use F1-score in addition to accuracy is that it is a more cautious measure when a dataset contains class imbalance.

The labels are set globally according to the presence of pain induction from the recording protocol for the different video clips before frame extraction, meaning that frames belonging to the same clip always have the same ground truth. Nevertheless, the loss function is optimized per frame during training, which means that the network outputs one classification per frame. When computing the F1-score at evaluation time for sequential models, we want to measure their performance on the aggregate 10-frame sequence-level. To this end, a majority vote is taken of the classifications across the sequence. If there is a tie, the vote is random.

\begin{table}[b]
\vspace{-3mm}
\setlength{\tabcolsep}{6pt}
\centering
\caption{\small Overview of input details for the four evaluated models.
}
\vspace{-2mm}
\label{table:impl}
\begin{tabular}{@{}l|l|l|l|l@{}}
\toprule
\scriptsize Model    & \scriptsize Input, Equine data   
& \scriptsize  Batch size & \scriptsize Sequential & \scriptsize Optimizer \\ \hline
\scriptsize C-LSTM-1 & \scriptsize  {[}10, 128, 128, 3{]} & \scriptsize 16   & \scriptsize Yes & \scriptsize Adadelta    \\
\scriptsize C-LSTM-2 & \scriptsize [10, 128, 128, 3{]} $\times$2 & \scriptsize 8 & \scriptsize Yes & \scriptsize Adadelta \\
\scriptsize VGG+LSTM \cite{Rodriguez2017} & \scriptsize  {[}10, 128, 128, 3{]} & \scriptsize 16  & \scriptsize Yes  & \scriptsize Adadelta \\
\scriptsize VGG+LSTM-2 \cite{Rodriguez2017} & \scriptsize  {[}10, 128, 128, 3{]} $\times$2 & \scriptsize 8  & \scriptsize Yes  & \scriptsize Adadelta \\
\scriptsize InceptionV3 \cite{Szegedy2014} & \scriptsize  {[}320, 240, 3{]}
& \scriptsize 100  & \scriptsize No  & \scriptsize RMSProp   \\
\scriptsize InceptionV3-2  \cite{Szegedy2014}& \scriptsize  {[}320, 240, 3{] $\times$2}           
& \scriptsize 50  & \scriptsize No  & \scriptsize RMSProp   \\ \bottomrule
\end{tabular}
\end{table}

\subsection{Veterinary expert baseline experiment}

As a baseline comparison, four veterinarians with expert training in recognizing equine pain were engaged in classifying 51 five second-clips sampled at random from the dataset (but distributed across the subjects) as pain or no pain. Five seconds is the same temporal footprint that the sequential models are trained and tested on. The results in Table \ref{table:1streamresults} show that it is not evident, even for an expert, to perform this kind of classification, highlighting the challenging nature of the task and dataset. The average F1-score (accuracy) of the experts was $54.6$ ($58.0\%$).

Since the clips are randomly extracted from the videos, it can happen that the horse looks away or interacts too much with an on-site observer. When the experts judged that it was not possible to assess pain from a certain clip, they did not respond. Thus, those results are not part of the averages. On average 5-10 clips out of the 51 were discarded by every rater. This "cherry-picking" should speak to the advantage of the experts compared to the models, as well as the fact that the experts could re-watch the clips. Yet, on average, the experts perform 18.9 percentage points (p.p.) worse than the C-LSTM-2 model.

Even so, the F1-scores varied among the raters. Their individual results were $52.7\%$, $50.7\%$, $72.2\%$ and $42.7\%$. Interestingly, the rater that obtained $72.2\%$ knew the dataset best out of the four and was present throughout the data collection. This rater's result, close to that of our best model, can thus be seen, to some extent, as testing on training data.

\begin{table*}[]
\setlength{\tabcolsep}{10pt}
\centering
\caption{\small Results (\% F1-score and accuracy) for binary pain classification per evaluated model. C-LSTM-2 (add) has the best average result. InceptionV3 is a single-frame model, while VGG+LSTM and C-LSTM are sequential models. Data augmentation was only applied to C-LSTM-2 (add), for computational reasons, since it performed better than the one with multiplicative fusion.}
\label{table:1streamresults}
\vspace{-2mm}
\begin{tabular}{@{}l|ll|ll|r@{}}
\toprule
                       & \multicolumn{2}{l}{\small \textbf{No data augmentation}}                        & \multicolumn{2}{l}{\small \textbf{Data augmentation}}                        \\ \midrule
                     
                       \textbf{One-Stream Models}                & \small \textbf{Avg. F1}              & \small \textbf{Avg. accuracy} & \small \textbf{Avg. F1}           & \small \textbf{Avg. accuracy} & \small \textbf{\# Parameters} \\ \midrule


\small InceptionV3 \cite{Szegedy2014}, Flow     & \small $ 54.1 \pm 10.9  $ & \small $ 60.8 \pm 10.3  $          & \small $ 57.5 \pm 12.3  $        & \small $ 60.4 \pm 8.5 $    &  \small $22,852,898$     \\

\small InceptionV3 \cite{Szegedy2014}, RGB   & \small $62.6 \pm 19.4 $     & \small $ 66.5 \pm 16.8 $          & \small $ 68.1 \pm 15.8 $              & \small $ 68.8 \pm 14.2 $    &  \small $22,852,898$     \\
\small VGG+LSTM \cite{Rodriguez2017}, Flow & \small $65.3 \pm 16.6 $                 & \small $ 67.6 \pm 13.6 $   & \small $ 50.5 \pm 17.0 $ & \small $ 60.9 \pm 13.2  $     & \small $57,713,474$ \\

\small VGG+LSTM \cite{Rodriguez2017}, RGB  & \small $69.4 \pm 28.8$                  & \small $72.8 \pm 22.9$           & \small $63.8 \pm 26.9 $      & \small $ 70.3 \pm 19.8$   & \small $57,713,474$      \\


\small C-LSTM-1, Flow (Ours)           & \small  $69.8 \pm 10.3$        & \small   $70.8 \pm 9.4$            &  \small $ 59.5 \pm 11.6 $    & \small $ 64.1 \pm 9.8  $     &  \small $731,522$              \\

\small C-LSTM-1, RGB (Ours)             & \small $ 71.3 \pm 19.4 $                 & \small  $73.5 \pm 16.3 $          & \small $64.0 \pm 20.5$               & \small $68.7 \pm 14.9 $ &  \small $731,522$       \\ \midrule


 \textbf{Two-Stream Models}                        &            &  &          &  & \\ \midrule
\small InceptionV3-2 (add)      & \small $62.4 \pm 20.5$  & \small $66.3 \pm 16.7$        & \small $ 55.4 \pm 18.0$  & \small $59.7 \pm 15.3$     & \small $45,704,770$  \\

\small VGG+LSTM-2 (add) &       \small $69.4 \pm 29.7$           & \small $74.6 \pm 22.5$           & \small $63.9 \pm 26.4$          & \small $70.5 \pm 18.9 $  & \small $115,425,922$  \\

\small C-LSTM-2 (mult.) (Ours)               & \small $67.3 \pm 19.1 $     & \small $ 72.2 \pm 14.9$           & -                          & -                  & \small $1,458,946$   \\

\small C-LSTM-2 (add) (Ours)                    & \small $\mathbf{73.5 \pm 18.3}$                  & \small $ 75.4 \pm 14.1 $          & \small $ 66.5 \pm 20.6  $         & \small $68.8 \pm 13.8 $    & \small $1,458,946$     \\ \midrule
 \textbf{Veterinary expert}                & \small $54.6 \pm 11.0 $                 & \small $58.0 \pm 13.6$           & \small n/a              & \small n/a                   &  \small n/a \\
 \bottomrule
\end{tabular}
\vspace{-3mm}
\end{table*}

\subsection{Automatic detection in the Equine Pain Dataset}

Table \ref{table:1streamresults} shows average results and standard deviations for the different evaluated models. The best obtained F1-score (accuracy) is $73.5 \pm 18.3$\% ($75.4 \pm 14.1$\%) using the C-LSTM-2 architecture. Both C-LSTM-1 and C-LSTM-2 in their best settings outperform the relevant baselines.

\vspace{-4mm}
\paragraph{The effect of two streams and temporal dynamics.}
We see a clear improvement of $+2.2$ p.p. from the C-LSTM one-stream model to its two-stream version, in terms of F1-score. For the VGG+LSTM model, there is no improvement from one to two streams when no data augmentation is used, and very little improvement ($+0.1$ p.p.) when data augmentation is used. The two-stream version of InceptionV3 in effect performs worse than its RGB one-stream version.
When comparing the importance of the RGB and optical flow streams, respectively, for the three different models, we see that the flow alone performs better for the C-LSTM-1 ($69.8\%$ F1-score) than for the VGG+LSTM ($66.1\%$ F1-score) and the InceptionV3 ($54.1\%$ F1-score). Out of the three models, the performance gap between the modalities is the least for the C-LSTM-1. The above observations combined interestingly show that dynamics and sequentiality are more important to the C-LSTM model.

Furthermore, we see in Table \ref{table:1streamresults} that the C-LSTM models perform better than the single-frame model InceptionV3\hk{, which can be considered as a baseline}. Its best result is $68.1\%$ F1-score for one stream with data augmentation. As a reminder, the input resolution to InceptionV3 is $320\times240$ pixels, compared to the $128\times128$ pixels for the C-LSTM networks. This underlines the better performance of the latter, even with nearly five times less pixels and orders of magnitude less parameters (Table \ref{table:1streamresults}).

\vspace{-4mm}
\paragraph{The effect of data augmentation.}

We can observe in Table~\ref{table:1streamresults} that the C-LSTM and VGG+LSTM models trained without augmented data perform better than those trained with augmented data. For InceptionV3, the performance is better with data augmentation. A possible explanation for why the C-LSTM models do not learn the augmented data as well as the clean data is their small number of parameters, compared to the other models (Table \ref{table:1streamresults}). The fact that the C-LSTM models do not improve with data augmentation suggests that they are of the appropriate size for the dataset and have not overfit. The VGG-16 base model of VGG+LSTM has many parameters but we hypothesize that it is the single LSTM layer that might be too minimalistic for the augmented data in this case. In the future, when working on more varied datasets, it will likely become necessary to use a larger number of hidden units, in the interest of stronger network expressivity and generalizing ability. Nevertheless, it is reasonable to use a smaller number of hidden units for equine data than for example when training on a larger and more varied dataset such as UCF-101 \cite{UCF-101}. For our applications, the data will always reside on a smaller manifold than general activity recognition datasets since we restrict ourselves to horses in a box or hospital setting. Data augmentation was only applied to the C-LSTM-2 (add) model since it performed better than the one with multiplicative fusion. In addition, the computational cost in training time was considerable, especially with augmented data.

\vspace{-4mm}
\paragraph{Performance difference between subjects.} The performance of VGG+LSTM varies markedly between subjects (Table \ref{table:persubject}). On that account, it has the highest global standard deviation among the models. When inspecting its classification decisions for subjects $1$ and $2$, for which it often scored above $90\%$ F1-score, the model had sometimes identified hay in the corner or the barred background as being significative of the pain category.
The horses it performed the worst on, subjects $5$ and $6$, have screens as background in the dataset and thus not as many details to overfit to.

As shown in Table \ref{table:persubject}, the results are generally better for subjects $1$-$5$ compared to subject $6$. Had we not included subject $6$ in our final score, our best average would instead have been $80.9 \pm 8.6\%$ F1-score. This horse displayed atypical behavior consisting of extreme playfulness and interaction with the human observer, because it had not been trained to stand still like the other subjects (Section \ref{sect:noisydata}).

The accuracies in Table \ref{table:1streamresults} in general have lower standard deviations than the F1-scores. This is because the F1-score discredits results where the model exclusively chooses one class for a whole test round. This often happened when the classification failed, as was almost always the case for subject 6. The F1-scores for subject 6 are around $40\%$, whereas in terms of accuracy these results are instead around $50\%$, hence reducing the standard deviation gap.

\begin{table}[!b]
\vspace{-3mm}
\setlength{\tabcolsep}{2pt}
\centering
\caption{\small Per-subject F1-scores for C-LSTM-2 and VGG+LSTM. The distribution is more even for C-LSTM-2.}
\vspace{-2mm}
\label{table:persubject}
\begin{tabular}{@{}l|l|l|l|l|l|l@{}}
\toprule
\footnotesize \textbf{Subject:}           		 & \footnotesize 1  		  & \footnotesize 2  		 		 & \footnotesize 3    		& \footnotesize 4    					& \footnotesize 5    & \footnotesize 6 \\ \hline
\tiny C-LSTM-2    & \tiny $ 87.1 \pm 4.8 $  & \tiny $83.4 \pm 5.6 $ & \tiny $79.7 \pm 11.2 $  & \tiny $ 73.4 \pm 5.8$   & \tiny $ 76.5 \pm 11.5$  & \tiny $40.8 \pm 3.6 $   \\

\tiny VGG+LSTM   & \tiny $ 90.5 \pm 4.1 $  & \tiny $ 94.4 \pm 3.2 $ & \tiny $ 77.2 \pm 16.3 $ & \tiny $ 89.1 \pm 4.8 $   & \tiny $ 32.1 \pm 7.0 $  & \tiny $ 33.8 \pm 4.2 $    \\

\bottomrule
\end{tabular}
\end{table}


\begin{figure*}[t!]
\begin{subfigure}[t]{\linewidth}
\centering
\includegraphics[width=1\linewidth]{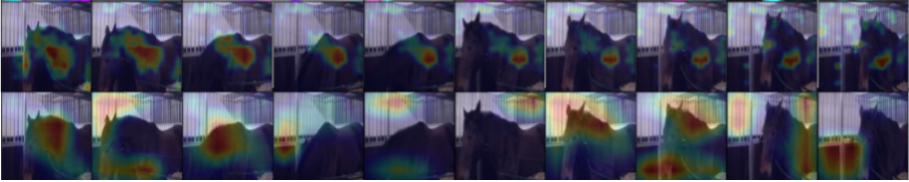}
\caption{\small A pain sequence correctly classified by both C-LSTM-2 and VGG+LSTM, \hknew{as well as by veterinarians}.}
\label{tp_pain}
\end{subfigure}
\begin{subfigure}[t]{\linewidth}
\centering
\includegraphics[width=1\linewidth]{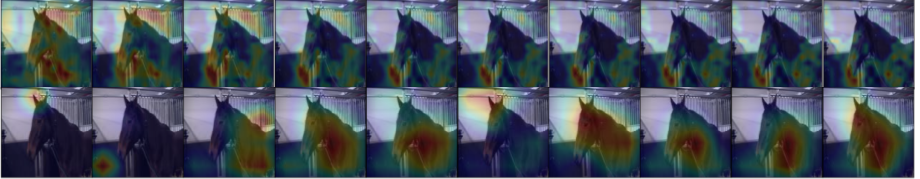}
\caption{\small A non-pain sequence correctly classified by both C-LSTM-2 and VGG+LSTM, misclassified by all veterinarians.}
\label{tn_nopain}
\end{subfigure}
\vspace{-2mm}
\caption{\small Saliency maps for two sequences and two different models (C-LSTM-2 on the first row, VGG+LSTM on the second row in each figure).  We observe that in general, the C-LSTM-2 model focuses on the horse and follows its movement, whereas the VGG+LSTM model has overfit (looks at the background) and does not follow a smooth temporal pattern across the sequence. In \ref{tp_pain}, the C-LSTM-2 is
attentive to the pose of the horses' shoulder and also zooms in on the tense eye area in the last three frames. In \ref{tn_nopain}, the muzzle, neck and eye area are in focus for the C-LSTM-2, whereas it appears that the VGG+LSTM has mostly learned to detect the presence of a horse.}
\vspace{-3mm}
\label{fig:gradcam}
\end{figure*}


\vspace{-4mm}
\paragraph{Tuning of C-LSTM-2.} For C-LSTM-2, the softer fusion by addition performs better than fusion by multiplication. The optical flow is computed at 16 fps and is matched to the corresponding 2 fps RGB frames. If the horse moves fast, there can be discrepancies when overlaying the RGB image with an AND mask (corresponding to multiplicative fusion) not fitting the shape of the horse. This could explain the better performance of the additive fusion.

\vspace{-4mm}
\paragraph{Comparison to Lu et al.}
\vspace{-1mm} 
\label{lucomparison}
To our knowledge, \cite{LuMahmoud} is the only previous study done on automatic pain recognition in larger non-human species.
An accuracy of 67\% is obtained on a three-class pain level classification task (76.8\% for two classes) via the detection of facial action units from images of sheep. The mapping from action units to pain level is done using the sheep pain facial expression scale, SPFES \cite{MCLENNAN201619}. However, this result is not obtained on a subject-exclusive train/test split and the data is culled for frontal faces with visible action units. Importantly, the labels are set based on action units, and not on induced pain. The fact that the ground truth is based on visible action units
makes the task easier. The authors also evaluate their method for unseen subjects; a confusion matrix is presented in \cite{LuMahmoud} with results from a three-class task. We translate this into two classes, to be able to compare our results. The F1-score (accuracy) when testing on unseen subjects is 58.4\% (58.1\%), 15 p.p. less than our best result.



\vspace{-4mm}
\paragraph{What do the models consider as pain?}
\label{gradcam}

We compute saliency maps for the two best models, C-LSTM-2 and VGG+LSTM, to investigate their classification decisions. Two example sequences are shown in Figure~\ref{fig:gradcam}.
The saliency maps are made using the Grad-CAM method 
\cite{Gradcam}, taking the gradient of the class output with respect to the different filters in the last convolutional layer (for C-LSTM-2, we take the gradient with respect to the last layer in the RGB-stream). The filters are then weighted according to their corresponding gradient magnitude. The one with the highest magnitude is considered the most critical for the classification decision. This particular filter, upsampled to the true image size, is then visualized as a heatmap.

The sequence in Figure~\ref{tn_nopain} shows a rather still horse with its head in an upright position. The sequence was misclassified as pain by all four experts but was correctly classified by the two computer models.
The veterinarians based their decision mostly on the tense triangular eye in the clip (sign of pain), visible in the proper resolution of the image, whereas the C-LSTM-2 seems to have mostly focused on the relaxed muzzle and upright head position (signs of non-pain). The VGG+LSTM seems to look at the entire horse, and frequently changes focus to the background and back again. Our impression is that the VGG+LSTM does not have a distinct enough focus to properly assess pain. Furthermore, Figure~\ref{tp_pain}, indicates that the VGG+LSTM model has overfit to the dataset, often looking at the background. The non-smooth temporal pattern of the saliency maps for VGG+LSTM in both Figures \ref{tp_pain} and \ref{tn_nopain} could be explained by this model's separation of spatial and temporal feature extraction. By contrast, C-LSTM-2 tends to follow relevant parts of the horse in a temporally smooth fashion, which gives us more confidence in its classifications. This study is not conclusive but gives an indication to what kind of patterns the models seem to learn from the data, as well as a preliminary explanation to the better performance of C-LSTM-2. Grad-CAM was applied to the single-frame model InceptionV3 as well, but the results were uninformative and showed no particular saliency, while at the same time the model gave over-confident classification decisions.


\section{Conclusions}
\label{sec:conc}
From our comparisons of three main models and their respective two-stream extensions, varying in the extent by which they model temporal dynamics, it appears that the spatio-temporal unfolding of behavior is crucial for pain recognition in horses. In this work, we have found that sequential imagery and a model such as the Convolutional LSTM that simultaneously processes the spatial and temporal features improves results compared to models that process single frames.
Sequentiality seems particularly important when assessing equine pain behavior, since horses do not convey a single straightforward face to show that they are in pain, the way humans can. Our obtained results on data labeled according to pain induction surpass a veterinary expert baseline and outperform the only previously reported result \cite{LuMahmoud} on subject-exclusive larger animal pain detection by a clear margin, using no facial action unit annotations or pre-processing of the data.


\subsection{Future work}

In future work, we would like to systematically investigate what the model finds salient when it comes to equine pain behavior and compare it to today's veterinary pain scoring. Interpretability is becoming increasingly important in deep learning, as shown by works such as \cite{Feichten18Visualize} and \cite{Gradcam}. We see an interesting research direction for exploring the spatio-temporal patterns of equine pain both in terms of the learned representations and classification attribution.
In the coming months, we will append more video labeled in terms of pain behavior to the database of \cite{Gleerup2015}.

{\small
\bibliographystyle{ieee}
\bibliography{refs}
}

\end{document}